\documentclass[10pt,twocolumn,letterpaper]{article}

\usepackage{cvpr}              

\definecolor{cvprblue}{rgb}{0.21,0.49,0.74}
\usepackage[pagebackref,breaklinks,colorlinks,allcolors=cvprblue]{hyperref}

\usepackage{algorithm}  
\usepackage{algorithmic}  

\usepackage{multirow}

\usepackage{multicol}

\usepackage{color}
\definecolor{green}{rgb}{0, 0.4, 0}
\definecolor{red}{rgb}{1.0, 0.0, 0.0}
\definecolor{teal}{rgb}{0.0, 0.4, 0.4}
\definecolor{purple}{rgb}{0.65,0,0.65}
\definecolor{saffron}{rgb}{0.95,0.75,0.2}
\definecolor{turquoise}{rgb}{0.0,0.5,0.5}
\definecolor{brown}{rgb}{0.5, 0.16, 0.16}
\definecolor{darkgreen}{rgb}{0, 0.75, 0}  

\usepackage{overpic}

\definecolor{lightgray}{rgb}{0.6, 0.6, 0.6}

\definecolor{DeltaColor}{rgb}{0.039,0.73,0.71}
\definecolor{SetaColor}{rgb}{0.867, 0.0235, 0.376}
\definecolor{SigmaColor}{rgb}{0.98,0.45,0.0}
\definecolor{RedColor}{rgb}{0.8,0,0}
\definecolor{AlphaColor}{rgb}{1.0, 0.4, 0.8}
\definecolor{BetaColor}{rgb}{0.8,0,0.8}
\definecolor{GammaColor}{rgb}{0.0,0,0.7}
\definecolor{EpsilonColor}{rgb}{0.353,0.725,0.906}
\definecolor{TauColor}{rgb}{0.423,0.235,0.192}

\def\paperID{2342} 
\def\confName{CVPR}
\def\confYear{2025}

\title{BoxFusion: Reconstruction-Free Open-Vocabulary 3D Object Detection via Real-Time Multi-View Box Fusion}

\author{
  Yuqing Lan\textsuperscript{1} \,
  Chenyang Zhu\textsuperscript{1}$^{\dagger}$ \,
  Zhirui Gao\textsuperscript{1} \,
  Jiazhao Zhang\textsuperscript{3} \,
  Yihan Cao\textsuperscript{1} \,\\  
  Renjiao Yi\textsuperscript{1} \,
  Yijie Wang\textsuperscript{1}$^{\dagger}$ \,
  Kai Xu\textsuperscript{1,2} \\  
  \textsuperscript{1}National University of Defense Technology \quad
  \textsuperscript{2}Xiangjiang Laboratory  \quad
  \textsuperscript{3}Peking University
}

\begin{document}
\maketitle

\renewcommand{\thefootnote}{}  
\footnote{$^{\dagger}$Corresponding authors}

\begin{abstract}
 
\definecolor{url_color}{rgb}{0.96, 0.22 , 0.6}
\hypersetup{
    colorlinks=true,       
    linkcolor=url_color,        
    citecolor=url_color,        
    urlcolor=url_color          
}

Open-vocabulary 3D object detection has gained significant interest due to its critical applications in autonomous driving and embodied AI. Existing detection methods, whether offline or online, typically rely on dense point cloud reconstruction, which imposes substantial computational overhead and memory constraints, hindering real-time deployment in downstream tasks. To address this, we propose a novel reconstruction-free online framework tailored for memory-efficient and real-time 3D detection. Specifically, given streaming posed RGB-D video input, we leverage Cubify Anything as a pre-trained visual foundation model (VFM) for single-view 3D object detection, coupled with CLIP to capture open-vocabulary semantics of detected objects. 
To fuse all detected bounding boxes across different views into a unified one, we employ an association module for correspondences of multi-views and an optimization module to fuse the 3D bounding boxes of the same instance.
The association module utilizes 3D Non-Maximum Suppression (NMS) and a box correspondence matching module. The optimization module uses an IoU-guided efficient random optimization technique based on particle filtering to enforce multi-view consistency of the 3D bounding boxes while minimizing computational complexity. 
Extensive experiments on CA-1M and ScanNetV2 datasets demonstrate that our method achieves state-of-the-art performance among online methods. Benefiting from this novel reconstruction-free paradigm for 3D object detection, our method exhibits great generalization abilities in various scenarios, enabling real-time perception even in environments exceeding 1000 square meters. 
The code will be released at \href{https://lanlan96.github.io/BoxFusion/}{project page}.

\end{abstract}    
\section{Introduction}

Recently, embodied intelligence has attracted great interest, 
with
many well-developed applications, such as embodied navigation~\cite{cao2024cognav,yin2024sg,zhou2024navgpt}, mobile manipulation~\cite{zhang2024gamma,yan2025m}. These embodied applications 
are mainly based
on scene understanding (e.g., 3D object detection~\cite{rukhovich2022fcaf3d, rukhovich2023tr3d}, semantic segmentation~\cite{jiang2024open, peng2024oa}, pose estimation~\cite{gao2025generic} and depth estimation~\cite{yang2024depth, lin2024promptda}). 
Current methods reconstruct the scene first for scene understanding with explicit representations, such as point clouds\cite{qi2019deep}, voxels\cite{zhou2018voxelnet}, etc. However, inspired by~\cite{li2017grass}, we found that a sparse representation of reality, which only preserves the position, scale, and semantics of objects' bounding boxes, as a scene representation, is sufficient to support the downstream embodied tasks, such as scene graph generation~\cite{koch2024open3dsg} and collision detection~\cite{wang2024appa}.

\begin{figure}[!t]
  \centering
  \includegraphics[width=0.45\textwidth]{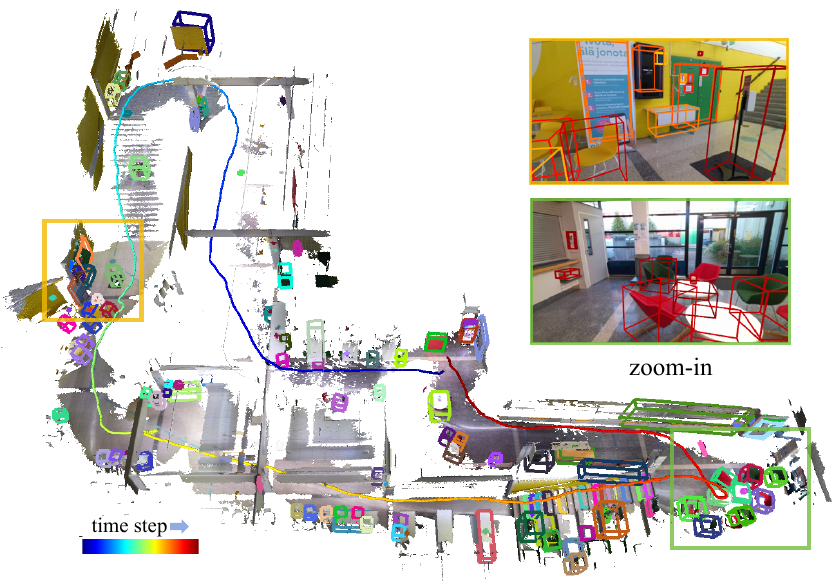} %
  \caption{The state-of-the art methods leverage dense reconstruction for online 3D perception. Our method is reconstruction-free, which can efficiently perform fine-grained 3D object detection with open-vocabulary semantics with over 20 FPS and 7GB GPU memory usage even in a multi-floor building (over 1000$m^2$). Note that the reconstructed mesh is only for visualization here.}
  \label{fig:teaser}
\end{figure}

Primary research in 3D object detection has focused predominantly on representations of 3D point clouds or meshes in the past. 
As inputs to these methods, point clouds inevitably undergo sparsification operations for computational feasibility, which introduce detection deficiency caused by lower resolution. 
Learning-based methods aim to incorporate inductive biases into models to mitigate this issue~\cite{qi2019deep,rukhovich2022fcaf3d,duan2022disarm}. Yet, the substantial discrepancy between 3D representations and physical reality restricts their applicability primarily to large, prominent objects. Therefore, some point-based methods~\cite{qi2020imvotenet, cao2023coda, cao2025collaborative} have attempted to use images as auxiliary inputs to enhance the performance.
Recently, with the development of visual foundation models (VFMs)~\cite{radford2021learning,kirillov2023segment,liu2024grounding}, researchers have started to leverage VFMs to incorporate rich semantic information from 2D images into point cloud-based methods, enabling 
open-vocabulary detections
~\cite{brazil2023omni3d,huang2024training,lu2023open}. 
While these methods are still dependent on dense point cloud reconstruction, 
which is prone to being affected by noise (e.g., accumulated errors in the reconstruction or noise in the sensors), or storage limitations of large-scale scenarios.

Furthermore, these methods typically follow a two-stage paradigm that requires complete point cloud reconstruction as inputs, which significantly hinder real-time applications. 
To address these challenges, some recent works have explored online perception methods. These methods either introduce a memory adapter~\cite{xu2024memory} to store the reconstructed point cloud or utilize 2D foundational models to extract fine-grained object masks and semantics from each view and then fuse them with the reconstructed dense point cloud for open-vocabulary perception~\cite{xu2024embodiedsam,tang2025onlineanyseg}. However, these online perception methods still rely on dense point-cloud reconstruction as input, which significantly increases computational overhead and degrades real-time performance, making it difficult to meet the requirements of real-time interaction in many downstream tasks.

To overcome these challenges, in this work, we propose a novel reconstruction-free online framework specifically designed for memory-efficient and real-time open-vocabulary 3D object detection. 
Our approach is inspired by many embodied tasks, such as navigation and manipulation, which do not require a complete reconstruction of the scene but rather focus on detecting and understanding the objects within it. As shown in Figure~\ref{fig:teaser}, by eliminating the need for dense point-cloud reconstruction, we can achieve faster and more efficient 3D object detection while maintaining high accuracy in large-scale scenarios.
Specifically, for streaming RGB-D video input, we leverage Cubify Anything~\cite{lazarow2024cubify} as a pre-trained visual foundation model (VFM) for single-view 3D object proposal generation. These proposals are enriched with open-vocabulary semantics by using CLIP~\cite{radford2021learning} to extract the semantic features of the detected objects.
An association module then aggregates multi-view proposals corresponding to the same instance through 3D Non-Maximum Suppression (NMS) and 2D box correspondence matching. Since the proposals from Cubify Anything are local and view-specific, it is of great necessity to aggregate and optimize the associated proposals from different views of the same object. Therefore, to further enhance multi-view consistency, we employ an optimization module, an IoU-guided efficient random optimization technique, which is grounded by particle filtering to overcome the highly non-linear optimization of object boxes. In this way, we can enforce multi-view consistency of the proposals of 3D bounding boxes while minimizing computational complexity to achieve real-time 3D object detection.

Extensive experiments on the CA-1M and ScanNetV2 datasets demonstrate that our method achieves state-of-the-art performance among online open-vocabulary 3D detection methods. Benefiting from the reconstruction-free paradigm, our approach exhibits exceptional generalization capabilities and enables real-time performance even in large-scale environments exceeding 1000 square meters, paving the way for the deployment in large-scale and challenging scenarios. In summary, our contributions are as follows:
\begin{itemize}
    \item We propose a novel reconstruction-free paradigm for online open-vocabulary 3D object detection, which models structural object layouts with desirable running and memory efficiency.
    \item We propose a multi-view box fusion technique based on particle filtering using random optimization, enabling real-time and consistent 3D bounding box detection.
    \item We have implemented an efficient system of online open-vocabulary 3D object detection. Extensive experiments validate the superior performance and robustness of our method in various challenging scenarios. 
\end{itemize}

\begin{figure*}[!t]
    \centering %
      \includegraphics[width=1.0\linewidth]{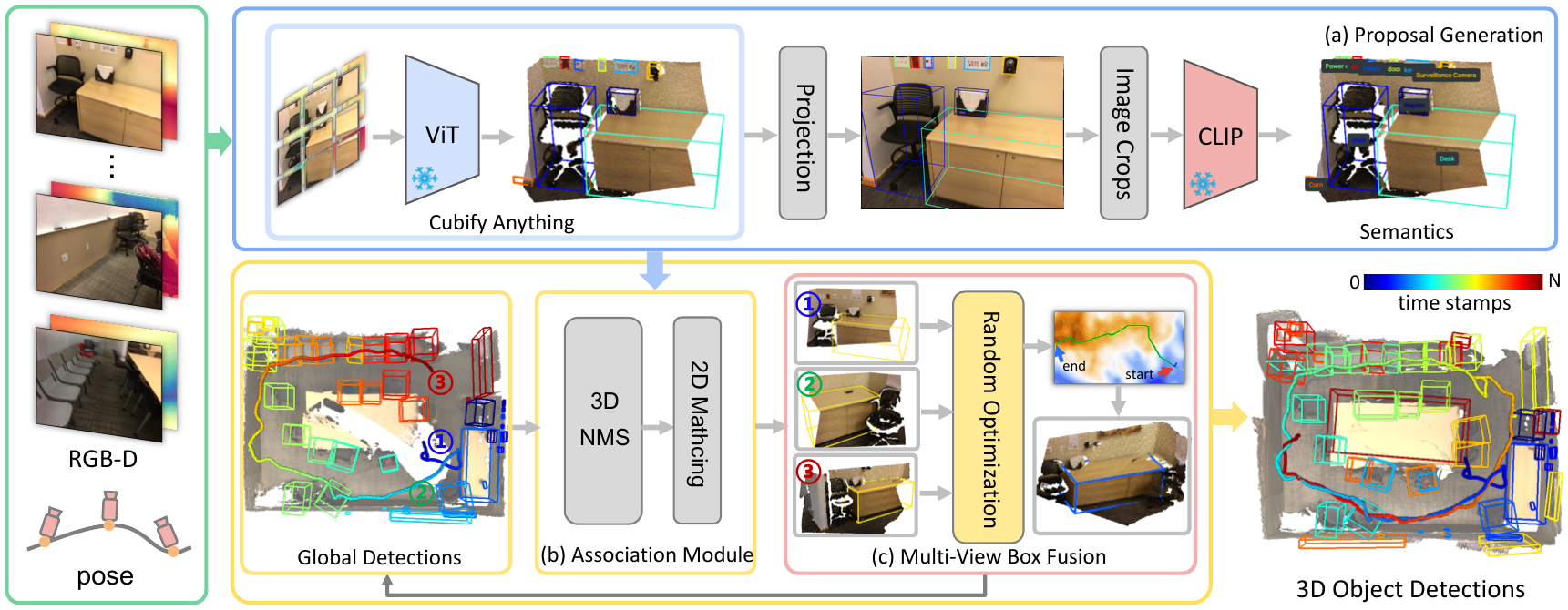} 
      \caption{Method Overview of BoxFusion. (a) Given online RGB-D images with camera poses, we use Cubify Anything to generate bounding box proposals for each keyframe, and project them to image planes to get the image crops and obtain open-vocabulary semantics with CLIP. (b) We employ an association module to perform 3D NMS for global boxes with proposals from the new keyframe, removing redundant boxes and associating those belonging to the same object. (c) With the associated candidate boxes of an object, we adopt random optimization based on particle filtering to fuse these candidate boxes into a multi-view consistent one using the IoU of the convex hulls of projected box corners. (e.g., the highlight desk is partially covered by three bounding boxes at three timestamps, which are not accurate enough). In this way, our method can efficiently detect fine-grained objects in real time without reconstruction.
      }
      \label{fig:pipeline}
\end{figure*}

\section{Related Work}

\textbf{Traditional 3D object detection:} 
Pioneered by PointNet~\cite{qi2017pointnet}, 3D object detection methods used to rely on dense labeled data corresponding to closed-set categories of objects. Point-based methods~\cite{qi2019deep,xie2020mlcvnet,liu2021group} follow the feature extraction techniques used in 2D object detection, and extract permutation-invariant point cloud features by 3D backbones, coupled with 3D detection heads. The follow-up works~\cite{engelmann20203d,chen2020hierarchical,lan2022arm3d} boost the performance by incorporating graph neural networks, better proposal generation, and contextual information. Voxel-based methods~\cite{zhou2018voxelnet,rukhovich2022fcaf3d,rukhovich2023tr3d}, akin to 2D detection, convert point clouds into regular grids, equipped with 3D convolutional neural networks (CNNs) for feature extraction. However, these methods focus on notable room-level objects (e.g., chairs and tables) and show poor generalization on objects of unseen categories, hindering the applications on more complicated tasks.

\noindent
\textbf{Open-vocabulary 3D object detection:}
With the development of visual foundation model (VFM) like SAM~\cite{kirillov2023segment} or DINO~\cite{liu2024grounding}, recent works~\cite{lu2023open,cao2023coda,wang2024ov,huang2024training} first establish the correspondence between 3D data like point clouds and 2D RGB images, and subsequently employ vision-language model (VLM) like CLIP~\cite{shafiullah2022clip} to segment images for more fruitful object proposals and corresponding semantics. These methods demonstrate improvements in detecting objects of novel categories in 3D scenes and make cross-modal retrieval between images and text descriptions feasible. However, these methods still rely on 3D data like point clouds as the core representation, suffering from the relatively poor resolution and noise of point clouds. This results in high computational overheads and limits the flexibility. SpatialLM~\cite{spatiallm} encodes the reconstructed point cloud and detects all the objects queried by a designed text prompt. Another paradigm is to leverage the 2D image features for 3D object detection, predicting the 3D bounding box directly. Cubify Anything~\cite{lazarow2024cubify} predicts metric fine-grained 3D bounding boxes by using a novel large-scale dataset and Encoder-Decoder vision transformer (ViT). DetAny3D~\cite{zhang2025detect} proposes a novel 3D detection framework that directly predicts 3D bounding boxes for a single image, leveraging the rich image features from VFM. However, these methods are primarily designed for single-view 3D object detection and exhibit limited generalization capabilities in scene-level online scanning scenarios.

\noindent
\textbf{Online 3D perception:}
Online 3D perception is a crucial task in robotics, enabling real-time scene understanding towards the surrounding environments. ~\cite{xu2024memory} employs a memory-based adapter for online 3D perception, including semantic and instance segmentation and 3D object detection. However, it is limited to closed-set categories and does not support open-vocabulary detection. Recent works~\cite{yang2023sam3d,xu2024embodiedsam,tang2025onlineanyseg} exploit the potential of 2D VFM like SAM~\cite{kirillov2023segment} or CropFormer~\cite{qi2023high} for online 3D instance segmentation via per-frame mask merging. 
However, multi-view fusion relying on 3D data representations (e.g., point clouds) inherently introduces sparsification artifacts and sensor-induced noise during reconstruction (e.g., depth holes from invalid measurements). Consequently, these methods, which utilize the spatial overlap of different masks for aggregation, are fragile, especially for small objects. The explicit reconstruction also increases the computational overhead and reduces the running and memory efficiency in terms of online settings and larger-scale scenarios. Compared to these methods, our method is reconstruction-free and more efficient in diverse scenarios.

\section{Methodology}

In this section, we first introduce the overall pipeline of our method. The insight of our method is to leverage foundation models to generate single-view fine-grained 3D bounding boxes and fuse them into multi-view consistent global 3D bounding boxes. In this way, the proposed reconstruction-free pipeline can efficiently generate scene-level 3D bounding boxes of fine-grained objects in real time. 

\subsection{Overview}
As shown in Figure~\ref{fig:pipeline}, our method consists of three main components: proposal generation, box association, and box fusion. Given RGB-D posed images as inputs, we first predict single-view and local 3D box proposals using Cubify Anything~\cite{lazarow2024cubify} for each keyframe, and transform the boxes from camera to world coordinates. These 3D proposals are then projected to 2D in order to crop the images and use CLIP~\cite{radford2021learning} for semantic features.
To associate the candidate boxes of the same instance, we first use 3D NMS to filter spatially overlapping boxes, coupled with a 2D correspondence matching module to associate small objects, since proposals of small objects are more likely to be adjacent but not close enough to have overlaps. Finally, we fuse these proposals into a single global 3D bounding box for each object in the scene. Specifically, inspired by~\cite{zhang2021rosefusion,zhang2022asro}, we employ the IoU-guided random optimization equipped with a pre-sampled particle swarm templates (PST) for better online efficiency and robustness to the highly-nonlinear optimization of box fusion.
In this way, the proposed method is robust and efficient for online open-vocabulary 3D object detection.

\subsection{Proposal Generation}
\label{sec:3-1}
In the era of foundation models, 3D object detection has been significantly boosted by leveraging large-scale pre-trained models. In this work, we adopt Cubify Anything~\cite{lazarow2024cubify} for fine-grained box proposal generation. Cubify Anything is a state-of-the-art method that can generate 3D bounding boxes from a single RGB-D image, and it is trained on a large-scale dataset with exhaustive objects that are labeled with ground-truth boxes. 

To be specific, given the successive RGB-D inputs $\{(I_t, D_t)\}_{t=1}^N$ with $N$ images in total, where $I_t$ is the RGB frame and $D_t$ is the depth frame, we first obtain $M$ 3D bounding box proposals $\{{B}^i_t\}_{i=1}^M$ for each image using Cubify Anything. Each box ${B}^i_t$ is represented in the camera coordinate system. To transform the boxes to the world coordinate system, we apply the camera pose transformation $\mathcal{T}_t$ to each box ${B}^i_t$, and obtain the global 3D boxes $\{\mathcal{G}^i_t\}_{i=1}^M$. In order to obtain the boxes with the corresponding semantics, we project the 3D boxes back to the 2D image planes $I_t$ and crop the corresponding image regions according to the projected box corners. The cropped images are then fed into a pre-trained CLIP model $\Theta$~\cite{radford2021learning} to obtain the semantic features $\{F_i\}_{i=1}^M$ of each box proposal as shown in Eq. \ref{eq:clip}.
\begin{equation}
    \label{eq:clip}
    F^i_t = \Theta( P(I_t, K_t, \{{B}^i_t\})),
\end{equation}
where $K_t$ denotes the camera intrinsics and $P$ indicates the projection and image cropping operation. $i$ and $t$ denote the box index and frame index.

\subsection{Association Module}
In this section, we introduce the box association module to associate the 3D bounding boxes of the same instance across different views. The association is crucial for maintaining consistent global 3D bounding boxes since numerous bounding boxes are generated from each single view. Previous methods~\cite{xu2024embodiedsam,tang2025onlineanyseg} mainly exploit spatial overlap for association, which is valid for most room-level common objects like chairs and desks, but can not generalize to small objects (e.g., a TV remote on the table). The reason is that proposals for most small objects generated from different methods are usually adjacent but not spatially overlapped, which may result in the missing association and detection.

To address this challenge, we propose a novel two-stage association module: 3D Non-Maximum Suppression (NMS) for associating spatially overlapping boxes, and a 2D correspondence matching module for associating the remaining fine-grained objects' boxes. The spatial association and correspondence association are introduced as follows.

\begin{figure}[!t]
    \centering
    \includegraphics[width=0.48\textwidth]{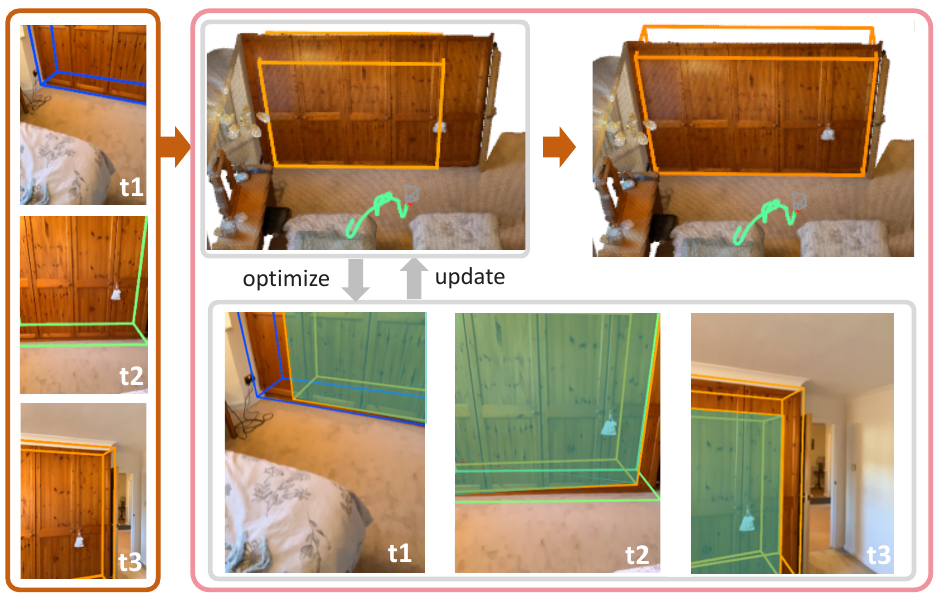} 
    \caption{Visualization of multi-view box fusion using random optimization with the IoU of the intersecting convex hulls of projected corners between the global box and single-view proposals. Three proposals of a cabinet (bottom) are fused into a global one (top).}
    \label{fig:iou-example}
    \vspace{-10pt}
\end{figure}

\subsubsection{Spatial Association}

Concretely, given the local box proposals $B_t$, we add these to global bounding boxes $\mathcal{G}_{t-1}$ using the camera pose and derive temporally current global boxes $\mathcal{G}_t$ at frame $t$. We need to determine whether any of the global boxes correspond to each other and remove the redundant ones. Compared to the axis-aligned NMS used in previous methods~\cite{qi2019deep,liu2021group}, we aim to generate scene-level oriented 3D bounding boxes. Therefore, we adopt the NMS for oriented bounding boxes. To be specific, we first sort $\mathcal{G}_t$ by the corresponding scores from Section~\ref{sec:3-1}. For each box $\mathcal{G}^i_t$ indexed by the descending order of scores, we calculate the IoU between $\mathcal{G}^i_t$ and the other boxes in $\mathcal{G}_t$. The IoU is calculated based on the convex hull of the 3D bounding boxes, which is more suitable for oriented boxes.

For online efficiency, we uniformly sample $O_n$ points in the convex hull of each pair of 3D bounding boxes, and only those that are likely to have overlap are considered. At first, this can be verified by checking whether the points on the box edges lie with each other. For brevity, we omit the time stamp $t$ in the following. For a pair of 3D bounding boxes $\mathcal{G}_i$ and $\mathcal{G}_j$ ($i>j$ in the descending order of scores), the 3D convex-based IoU $\Omega_{ij}$ is defined as Eq. \ref{eq:iou}. We manage a candidate list $\Psi_i$ for $\mathcal{G}_i$. We filter $\mathcal{G}_j$ if $\Omega_{ij}$ is higher than the threshold $\tau_{3d}$ and $\mathcal{G}_j$ is then associated with $\mathcal{G}_i$ if the camera direction is larger than $\tau_r$ or the camera translation is larger than $\tau_t$ compared to all candidate boxes in $\Psi_i$.

\begin{equation}
    \label{eq:iou}
    \Omega_{ij} = \frac{\sum_{k=1}^{O_n} \mathcal{C}(O_k, \mathcal{G}_i\cup\mathcal{G}_j )}{\sum_{k=1}^{O_n} \mathcal{C}(O_k, \mathcal{G}_i)+\sum_{k=1}^{O_n} \mathcal{C}(O_k, \mathcal{G}_j)},
\end{equation}
where $\mathcal{C}$ is the indicator function that returns 1 if the point $O_k \in \mathcal{G}_i$, otherwise 0 and $\mathcal{G}_i\cup\mathcal{G}_j$ denotes the union of convex hulls of 3D bounding box $\mathcal{G}_i$ and $\mathcal{G}_j$.

\subsubsection{Correspondence Association}
While the spatial association can reliably associate room-level common objects, it is challenging for small objects. Proposals for the same small object are typically less accurate and likely to be spatially adjacent but not overlapping. Inspired by~\cite{wang2024dust3r,leroy2024grounding}, we argue that image-based correspondences are efficient and robust to match box proposals from different camera views that belong to the same object instance. However, directly using pre-trained models to find correspondence between the current image $I_t$ and all historical frames that have visibility overlap is time-consuming and computationally expensive. Therefore, we propose a simple yet effective solution that utilizes 2D projective IoU to associate the remaining small objects. Taking the local object box $B_i$ in frame $t$ as an example, global 3D boxes $\mathcal{G}_{t-1}$ at frame $t-1$ are projected to the current image $I_t$. We check whether the local candidate box $\mathcal{B}_i$ corresponds to any global box in $\mathcal{G}_{t-1}$ with $K$ boxes by the IoU of projective bounding boxes as shown in Eq. \ref{eq:2d-cor}.
\begin{equation}
    \label{eq:2d-cor}
    j_{max}=arg\,max(\Phi(\mathcal{P}(T^{-1}_t, {B}_i),\mathcal{P} (T^{-1}_t,\mathcal{G}^j_t))), j\in K,
\end{equation}
where $\Phi$ denotes the 2D IoU of two convex hulls obtained from the projective corners of 3D bounding boxes. $\mathcal{P}$ refers to the projection using the pinhole camera assumption. If the IoU $\Phi_{i{j_{max}}}$ between the current candidate $B_i$ and global box $\mathcal{G}^{j_{max}}_t$ is larger than $\tau_{2d}$, similar strategies in the spatial association are leveraged to determine whether $B_i$ should be added to the candidate list $\Psi^{j_{max}}_t$ of $\mathcal{G}^{j_{max}}_t$ for further multi-view box fusion.

\subsection{Multi-view Box Fusion}
Since the proposals from different camera views are local and not entirely accurate, it is of great necessity to fuse these candidate boxes belonging to the same object into a multi-view consistent bounding box. For example, the sequential camera views only cover parts of a big cabinet, and there would be several separate 3D boxes for the cabinet without box fusion. 

As shown in Figure~\ref{fig:iou-example}, for global 3D boxes $\mathcal{G}^i_t$ and corresponding candidate box list $\Psi^i_t$, the goal is to aggregate the associated 3D boxes in $\Psi^i_t$ and fuse them into a global one that is consistent in multiple views. Empirically, we find the box proposals generated by foundation models like Cubify Anything~\cite{lazarow2024cubify} look accurate in 2D images but are biased from the ground truth in 3D with scale uncertainty. Therefore, we propose to optimize the position and scales of the global box $\mathcal{G}^i_t$ to fit the proposals $\Psi^i_t$ in different views of the same object. Given the position $\mathbf{p}=(x,y,z)$ and shape $\mathbf{s}=(l,w,h)$ of $\mathcal{G}^i_t$, which are initialized as the average of those of boxes in $\Psi^i_t$, we aim to optimize the 6D variables $(\mathbf{p},\mathbf{s})=(x,y,z,l,w,h)$ to enable the global box $\mathcal{G}^i_t$ to exactly cover each candidate box in the corresponding views.

\begin{figure}[!t]
    \centering
    \includegraphics[width=0.46\textwidth]{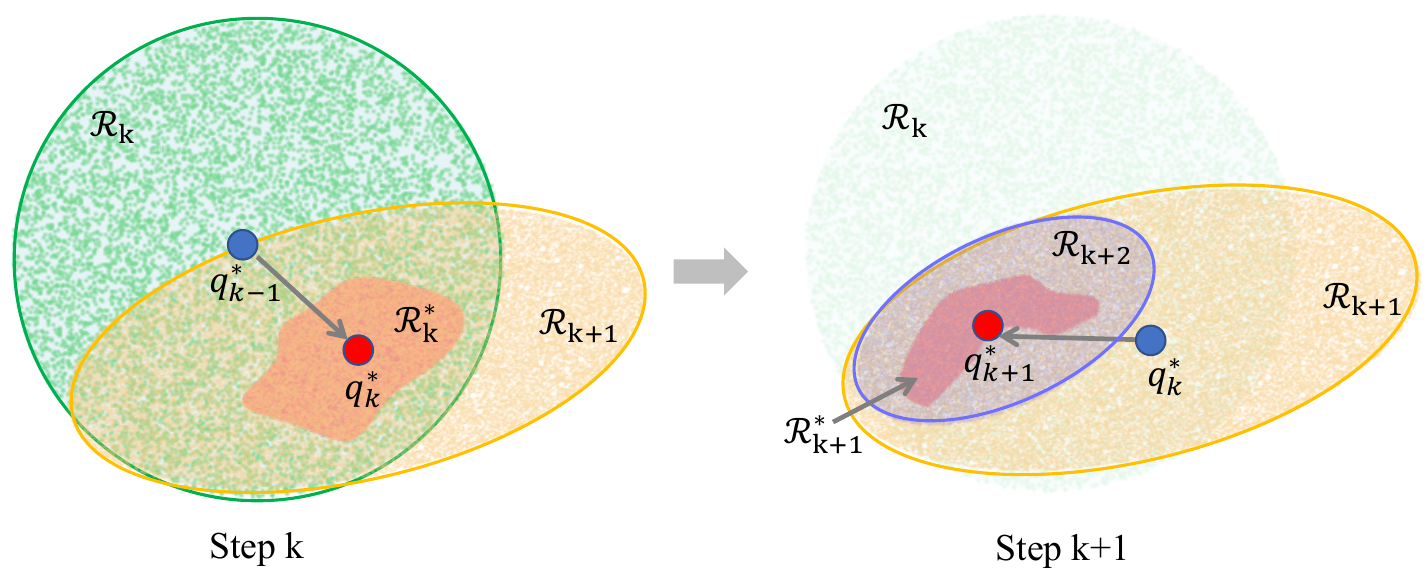} 
    \caption{2D example of particle filtering optimization (PFO) using pre-sampled swarm template (PST) for box fusion. At step $k$, particles in $\mathcal{R}^*_k$ are selected as the superior set, moving the center $q^*_{k-1}$ of particles $\mathcal{R}_k$ to $q^*_{k}$ and rescaling $\mathcal{R}_k$ to $\mathcal{R}_{k+1}$. Similar moving and rescaling is employed to the particle set $\mathcal{R}_{k+1}$ at step $k+1$. Therefore, the optimization can converge efficiently and accurately. }
    \label{fig:PST}
\end{figure}

Let us define the objective function first. The global box $\mathcal{G}^i_t$ with parametes $(\mathbf{p},\mathbf{s})$ to be optimized is projected to images $\{I_j| j\in \varphi^i_t\}$ with camera poses $\{T_j| j\in \varphi^i_t\}$, where $\varphi^i_t$ is the set of camera indices of $\Psi^i_t$. The projected box corners of the global box $\mathcal{G}^i_t$ in image $I_j$ are used to derive a 2D convex hull denoted as $\alpha_{ij}$. Similarly, we can obtain the 2D convex hull $\beta_{ij}$ of the candidate box ($j \in \varphi^i_t$). The goal is to maximize the 2D IoU between $\alpha_{ij}$ and $\beta_{ij}$ in each view. $\alpha_{ij}$ can be obtained by projecting the corners of $\mathcal{G}^i_t$ to the image $I_j$ using the camera pose $T_j$ and $\mathcal{Q}$ for the computation of convex hulls as Eq. \ref{eq:ro-proj}. The optimization can be formulated as follows:

\begin{equation}
    \label{eq:ro-proj}
    \alpha_{ij}=\mathcal{Q} (\mathcal{P}(T^{-1}_t,(\mathcal{G}^i_t(\mathbf{p},\mathbf{s})))).
\end{equation}

\begin{equation}
    \label{eq:ro-objective}
    (\mathbf{p}^{*},\mathbf{s}^{*})=\arg\max_{\mathbf{p},\mathbf{s}}\sum_{j\in \varphi^i_t}\mathcal{H}(\alpha_{ij},\beta_{ij}),
\end{equation}
where $\mathcal{H}$ is the 2D IoU function between two convex hulls.

The goal is to optimize the objective function in Eq. \ref{eq:ro-objective}, but the optimization is highly non-linear and challenging to solve in real time. Inspired by~\cite{ji2008particle,zhang2021rosefusion,tang2023mips,lan2025remixfusion}, to address this, we propose to use a \textit{pre-sampled particle swarm template (PST) }to guide the optimization. This process is based on \textit{particle filtering optimization (PFO)}, and the likelihood function corresponding to Eq. \ref{eq:ro-objective} can be formulated as \ref{eq:ro-pfo}.

\begin{equation}
    \label{eq:ro-pfo}
p(g|\mathbf{p},\mathbf{s})=\exp\left(-\frac{1}{\xi }\sum_{j\in \varphi^i_t}\mathcal{H}(\alpha_{ij},\beta_{ij})\right),
\end{equation}
where $g$ denotes the observation function and $\xi$ is a temperature.

The PST is a set of $N_{pst}$ pre-sampled particles uniformly sampled in 6D space corresponding to the position $\mathbf{p}=(x,y,z), \mathbf{p}\in[-1,1]$ and shape $\mathbf{s}=(l,w,h), \mathbf{s}\in[-1,1]$. With the initialized global box $\mathcal{G}^i_t(\mathbf{p}',\mathbf{s}')$, we apply the PST to $\mathcal{G}^i_t(\mathbf{p}',\mathbf{s}')$ and detect which particle fits better using the objective function in Eq. \ref{eq:ro-objective}. As shown in Figure~\ref{fig:PST}, we select the best particles in the PST $\mathcal{R}_k$ as a superior set $\mathcal{R}^*_k$, which is moved and rescaled according to the superior set $\mathcal{R}^*_k$ in step $k$. Then we can obtain the best transformation $q^*_k=(\mathbf{p}^*_k,\mathbf{s}^*_k)$, which are added to the current observation: $q'=(\mathbf{p}',\mathbf{s}')+(\mathbf{p}^*_k,\mathbf{s}^*_k)$. Once the iteration reaches the maximum number or the optimization converges, we use the current observation $(\mathbf{p}',\mathbf{s}')$ as the optimized one $(\mathbf{p}^*,\mathbf{s}^*)$. Subsequently, we update the global box $\mathcal{G}^i_t$ with the finally optimized parameters $(\mathbf{p}^*,\mathbf{s}^*)$. The proposed random optimization is performed once the number of candidate boxes $\Psi^i_t$ is larger than a threshold $\tau_{box}$, or new candidates are added.

\section{Experiments}

\begin{table*}[!t]

\centering
\setlength{\tabcolsep}{7.5pt}  
\renewcommand{\arraystretch}{1.4}  

\begin{tabular}{l|cc|ccc|ccc|c}
\specialrule{1.0pt}{0pt}{0pt}  
\multirow{2}*{\textbf{Method}} & \multirow{2}*{\textbf{Online}} & \multirow{2}*{\textbf{Open-Vocabulary}} & \multicolumn{3}{c|}{\textbf{CA-1M}} & \multicolumn{3}{c|}{\textbf{ScanNetV2}} & \multirow{2}*{\textbf{FPS}} \\
~ & ~ & ~ & $AP_{15}$ & $AP_{25}$ & $AP_{50}$ & $AP_{15}$ & $AP_{25}$ & $AP_{50}$ &  \\
\specialrule{1.0pt}{0pt}{0pt}
FCAF~\cite{rukhovich2022fcaf3d} & $\times$ & $\times$ & 6.39&5.25&2.74 & \textbf{66.64}&\textbf{66.1}&\textbf{53.31}& - \\
TR3D~\cite{rukhovich2023tr3d} & $\times$ & $\times$ &5.16&	4.32&2.41& 60.08&57.96&52.26  & - \\
SpatialLM~\cite{spatiallm} & $\times$ & \checkmark & 1.40 &0.78 &0.15 & 8.06& 4.04& 0.58  & - \\ 
\hline
EmbodiedSAM~\cite{xu2024embodiedsam} & \checkmark & $\times$ & 9.17&5.01&0.80 & 5.22 &2.46 &0.33  & 10 \\
OnlineAnySeg~\cite{tang2025onlineanyseg} & \checkmark & \checkmark & 6.93&5.19&1.79&31.39&	21.81&9.92  & 15 \\
\hline
Ours  & \checkmark & \checkmark & \textbf{31.22}& \textbf{25.66} & \textbf{8.75} & 37.46&31.36&	13.41   & \textbf{20} \\
\specialrule{1.0pt}{0pt}{0pt}  
\end{tabular}

\caption{Class-agnostic detection comparison on CA-1M and ScanNetV2 dataset. Point-based methods are not comparable on CA-1M, which can be attributed to closed-set training datasets and worse generalization to various objects. Our method is superior to both offline and online methods in CA-1M and performs the best on ScanNetV2 among the online methods. While scanNetV2 only provides ground-truth labels for common objects (18 categories), open-vocabulary methods detect many more objects, resulting in lower performance compared to offline point-based methods. Note that OnlineAnySeg segments images in an offline manner, and FPS is reported for the online fusion.}

\label{tab:whole_seq}
\end{table*}

\subsection{Experimental Setup}
\textbf{Datasets:} We evaluate our method on the CA-1M~\cite{lazarow2024cubify} and ScanNetV2~\cite{dai2017scannet} dataset. CA-1M is a large-scale dataset for class-agnostic 3D object detection with about 15M frames, including over 440K objects, which include lots of small but common objects in daily life. The ground truth 3D bounding boxes are labeled on the scan mesh from a FARO laser. There are 107 validation scenes in CA-1M. ScanNetV2 is a large RGB-D indoor dataset with a training set of 1201 scenes and a validation set of 312 scenes, which labels 18 categories of room-level objects.

\noindent
\textbf{Baselines:} We mainly compare our methods to state-of-the-art offline point-cloud based methods: FCAF~\cite{rukhovich2022fcaf3d}, TR3D~\cite{rukhovich2023tr3d} and SpatialLM~\cite{spatiallm}. Since there are no online open-vocabulary 3D detection alternatives, we take online perception methods: EmbodiedSAM~\cite{xu2024embodiedsam} and OnlineAnySeg~\cite{tang2025onlineanyseg} into consideration, which are initially designed for instance segmentation and are post-processed for gravity-aware 3D bounding box detection following~\cite{dhamo2021graph}.

\noindent
\textbf{Metrics:}
We use the standard metrics for 3D object detection following~\cite {qi2019deep,brazil2023omni3d}, including Average Precision (AP) at different IoU thresholds. For CA-1M and ScanNetV2, we report class-agnostic $AP$ at IoU thresholds of 0.15, 0.25, and 0.5. ScanNetV2 only provides ground-truth axis-aligned bounding boxes, which requires results of all methods to be axis-aligned for fair comparison. The running efficiency of the system FPS and GPU memory usage are considered as well.

\noindent
\textbf{Implementation Details:}
We evaluate BoxFusion on a desktop PC equipped with a 3.90GHz Intel Core i9-14900K CPU and an NVIDIA RTX 3090 Ti GPU. We implement the IoU-based random optimization based on the PyCUDA~\cite{klockner2012pycuda} libraries for acceleration. For the semantics of open-vocabulary categories, we use the text embedding of the provided categories (over 300) in~\cite{lin2014microsoft} to match the detected objects. More details can be found in the supplementary materials. We highly recommend referring to supplementary videos for a more intuitive understanding of our method.

\begin{figure*}[!t]
  \centering %
    \includegraphics[width=1.0\linewidth]{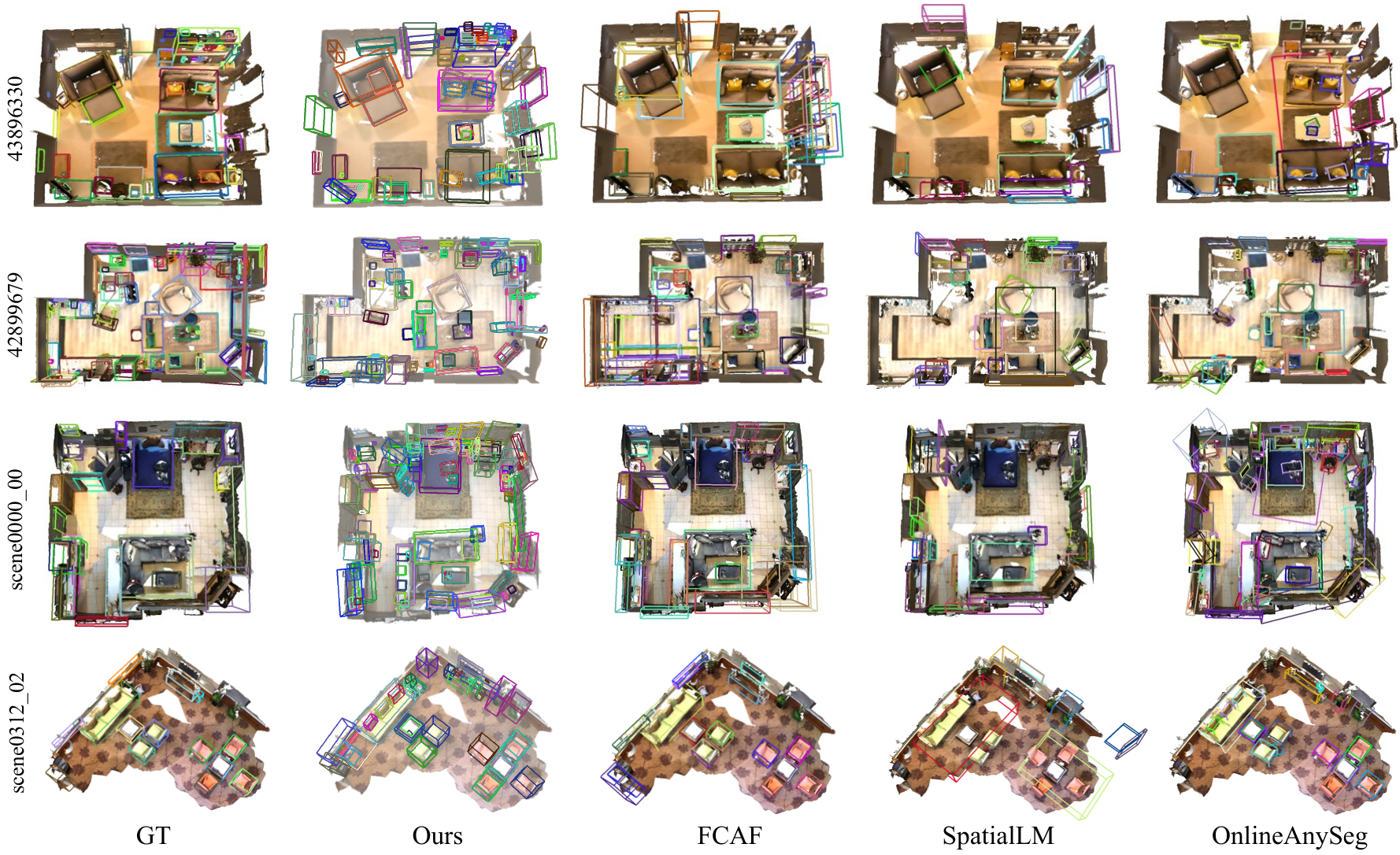}
    \caption{Gallery of 3D object detection on CA-1M and ScanNetV2. Our method is reconstruction-free, and the mesh is semi-transparent. Our method can accurately detect both the common objects and the objects that are not included in most closed-set datasets. Offline methods show poor generalization on fine-grained objects, and OnlineAnySeg (represented as the online method) is slightly better than offline methods but still struggles to detect fine-grained objects. Best viewed on screen.}
    \label{fig:detection}
\end{figure*}

\begin{figure*}[!t]
  \centering 
    \includegraphics[width=1.0\linewidth]{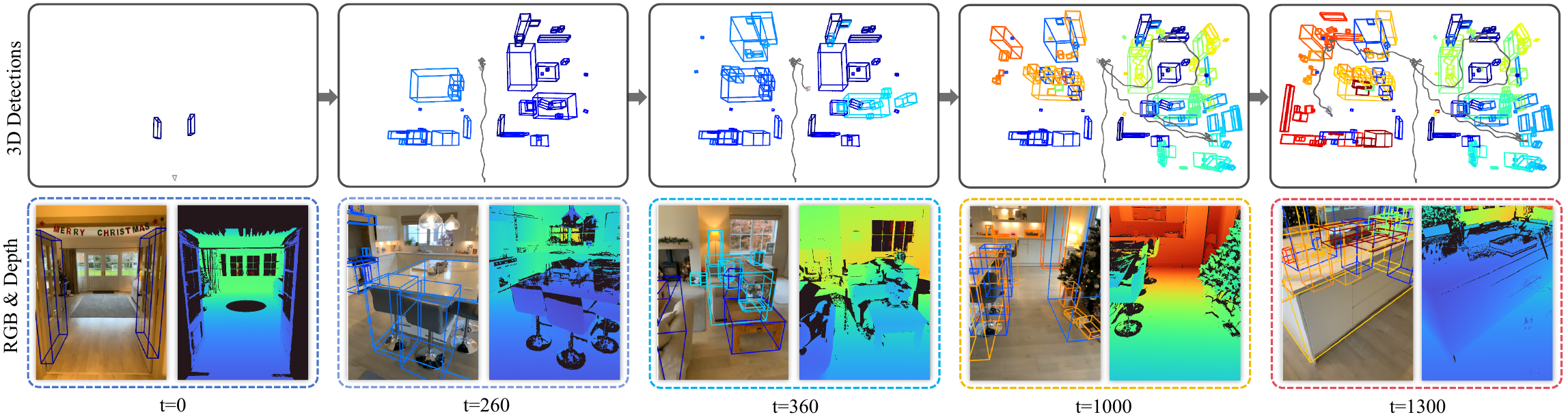}
    \caption{Visualization of online 3D object detections on 42898849 of CA-1M. The first row shows the detected bounding boxes alongside the camera trajectory (gray), with boxes colored from blue to red to indicate temporal order. The second row displays the RGB images with projected 3D boxes and corresponding depth images at different timestamps.}
    \label{fig:online}
\end{figure*}

\subsection{Quantitative Results}
The quantitative comparison of our method and state-of-the-art baselines on CA-1M and ScanNetV2 is shown in Table~\ref{tab:whole_seq}. All sequences of the CA-1M validation set and 100 uniformly sampled sequences of the ScanNetV2 validation set are considered. Class-agnostic detection is adopted, since it is easier to obtain the categories of objects with the help of visual foundation models. Our method achieves the best performance on both datasets among the online methods, especially on CA-1M, where our method outperforms the second-best method by a large margin ($22.02$ improvement in $AP_{15}$). Point-based methods~\cite{rukhovich2022fcaf3d,rukhovich2023tr3d,spatiallm} demonstrate poor generalization on CA-1M, which is significantly lower than online alternatives in terms of $AP_{15}$ to $AP_{50}$. The main reason is that point-based methods are trained on closed-set datasets, which focus on room-level object detection and typically ignore other ordinary and daily objects. 

Online methods like EmbodiedSAM~\cite{xu2024embodiedsam} and OnlineAnySeg~\cite{tang2025onlineanyseg} are designed for instance segmentation, and the results are transformed into bounding boxes for evaluation, following~\cite{lee2022patchworkpp}. Details about the transformation are demonstrated in the supplementary materials. Online methods aim to detect objects with streaming RGB-D inputs, which overall show better performance than offline methods. This may be attributed to that all online methods in Table~\ref{tab:whole_seq} can efficiently capture the rich and useful contexts in images with VFMs for more comprehensive scene understanding. 

Point-based method~\cite{rukhovich2022fcaf3d,rukhovich2023tr3d} outperforms the others on ScanNetV2. These methods are trained on closed-set datasets and struggle to detect objects of other common categories, which is discussed in Section~\ref{sec:quality}. $AP_{15}$ of our method is over 29 on both CA-1M and ScanNetV2, demonstrating the robustness of our method. 
Moreover, the significant improvement of $AP_{25}$ (20.32 and 6.3) on CA-1M and ScanNetV2 compared to the best online method illustrates that our method can detect objects effectively and accurately without the need for dense reconstruction, which is a significant advantage over other methods. Note that the FPS of OnlineAnySeg~\cite{tang2025onlineanyseg} is only for the fusion stage, and the segmentation of images is offline, which is time-consuming. Our method is over 20 FPS on average, which can detect and obtain the open-vocabulary semantic features or categories in real time.


\subsection{Qualitative Results}
\label{sec:quality}
Figure~\ref{fig:online} presents the visualization of online scanning of BoxFusion. The first row is the detected object boxes, which are colored from blue to red, indicating the temporal order. The second row displays the RGB image with 3D boxes projected into the image planes and the depth image, corresponding to the gray camera trajectory in the first row. Results show that our method can detect fine-grained objects in real time accurately and robustly.

As shown in Figure~\ref{fig:detection}, we compare the detections of our methods to other state-of-the-art baselines. The background of our method is semi-transparent, since we do not rely on reconstruction for detection. In the first two rows, the comparison of two sequences of the CA-1M dataset is displayed, where the ground-truth is more fine-grained than ScanNetV2. Our method stands out as the best with detailed and accurate method.  FCAF~\cite{rukhovich2022fcaf3d} is trained on closed-set datasets, which can detect the common objects (e.g., sofas and tables) accurately but struggle to detect the fine-grained objects (e.g., newspapers on the coffee table). SpatialLM~\cite{spatiallm} can only detect the big furniture accurately, and there are many missing objects in the detection. OnlineAnySeg~\cite{tang2025onlineanyseg} can successfully detect both the common objects and previously unseen objects thanks to the VFM. However, it is hard for OnlineAnySeg to detect the small but important objects (e.g., books and keys at the upper right of the first row). This can be attributed to the merging strategies that rely on point clouds for spatial overlaps, which may fail due to noise or errors in the reconstruction. In contrast, our method is reconstruction-free and leverages the rich context in multi-view images to detect and fuse the bounding boxes of 3D objects, leading to fine-grained and accurate 3D object detection.

\begin{figure*}[!t]
  \centering %
    \includegraphics[width=1.0\linewidth]{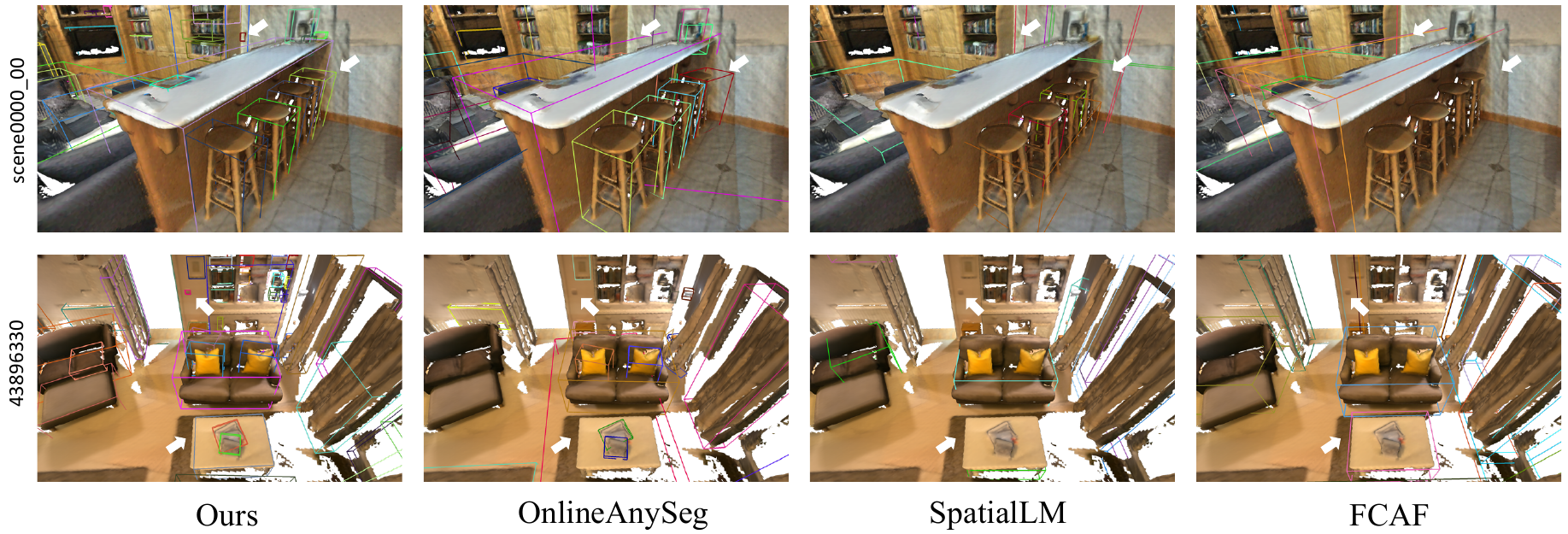}
    \caption{Detailed qualitative comparison on ScanNetV2 and CA-1M. Our method achieves comprehensive detection of both common objects and fine-grained entities (e.g., the occluded stool, the wall switch), while existing approaches exhibit limited perception accuracy and robustness. White arrows highlight the comparison.}
    \label{fig:zoomin}
\end{figure*}

The detailed qualitative comparison is demonstrated in Figure~\ref{fig:zoomin}. Our method preserves the most accurate and complete 3D detection, while other methods indicate inferior performance. The while arrows highlight the comparison. For example, our method can accurately detect the box of stool that is near the wall and partially occluded (the first row). The wall switch (the second row) is successfully detected and multi-view consistent for our method, while other methods fail to detect it.

\begin{figure}[htb]
    \centering
    \includegraphics[width=0.47\textwidth]{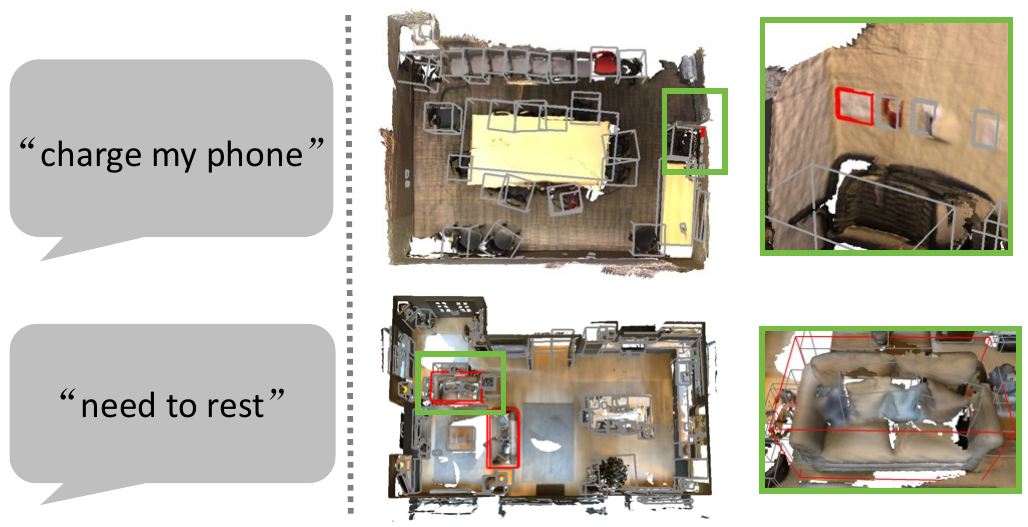} %
    \caption{Open-vocabulary object retrieval with user-specific text prompts. Mesh is only for visualization.}
    \label{fig:oa-query}
\end{figure}

As shown in Figure~\ref{fig:oa-query}, our method is open-vocabulary and supports object retrieval with user-specific text prompts. For example, if the prompt indicates it needs to charge the phone (the first row), the power outlet on the wall marked by the red rectangle is highlighted to correspond to the prompt. This indicates that our method can be easily adapted to many downstream embodied tasks.

\subsection{Ablation Studies}








\begin{table}[!t]

\centering
\setlength{\tabcolsep}{16pt}  
\renewcommand{\arraystretch}{1.2}  

\begin{tabular}{l|cc}
\specialrule{1.0pt}{0pt}{0pt}  
Method  & $AP_{15}$ & $AP_{25}$  \\
\specialrule{1.0pt}{0pt}{0pt}


No SA & 16.85&14.23  \\
No CA & 34.55&28.16 \\
No MVF & 29.03&25.17  \\
Ours & \textbf{35.49}& \textbf{28.31} \\

\specialrule{1.0pt}{0pt}{0pt}  
\end{tabular}

\caption{Ablation studies of different modules proposed in our method on six scenes of CA-1M and eight scenes of ScanNetV2.}
\label{tab:ablation}

\end{table}
We mainly evaluate the proposed modules, including the spatial association (SA for short), correspondence association (CA for short), and the multi-view box fusion (MVF for short). 
\begin{itemize}
    \item SA: Without the SA module, the $AP_{15}$ drops by 18.48, which indicates that the spatial association is important for global 3D object detection. It can provide spatially overlapping details to associate box proposals from different views. 
    \item CA: The box correspondence module is mainly leveraged to associate small objects, and is necessary to provide accurate association to benefit the multi-view box fusion.
    \item MVF: The proposed multi-view box fusion is utilized to generate consistent 3D bounding boxes based on the association modules. There is significant improvement in $AP_{15}$ and $AP_{25}$ (2.09 and 0.51 improvements) compared to not using the MVF module, which validates the effectiveness.
\end{itemize}

As shown in Table~\ref{tab:ablation}, the ablation studies are performed on CA-1M and ScanNetV2. The proposed modules are beneficial to each other. Therefore, the full method is able to efficiently associate the box proposals and fuse them into globally consistent boxes.

\begin{table}[!t]
  \centering
\caption{
Comparison of run-time system FPS and GPU memory usage 
(GPU mem. for short)
on the ScanNetV2 dataset. Our method (without dense reconstruction) is the most efficient and accurate online method.
}
\renewcommand{\arraystretch}{1.4}
\scalebox{1.0}{
\setlength{\tabcolsep}{1.15mm}{
\begin{tabular}{l|cccc}
    \toprule
        Method  & $AP_{15}$ $\uparrow$ & $AP_{25}$ $\uparrow$  & FPS $\uparrow$ & GPU mem. $\downarrow$ \\ \hline
        EmbodiedSAM & 5.22  & 2.46  & 10  & 12.5G    \\ 
        OnlineAnySeg & 31.39 & 21.81 & 15 &  21.1G   \\ 
        Ours & \textbf{37.46} & \textbf{31.36} & \textbf{20} & \textbf{7.0G} \\ 
        \bottomrule
\end{tabular}
}}
\label{tab:runtime}
\end{table} 

\begin{figure}[htb]
    \centering
    \includegraphics[width=0.47\textwidth]{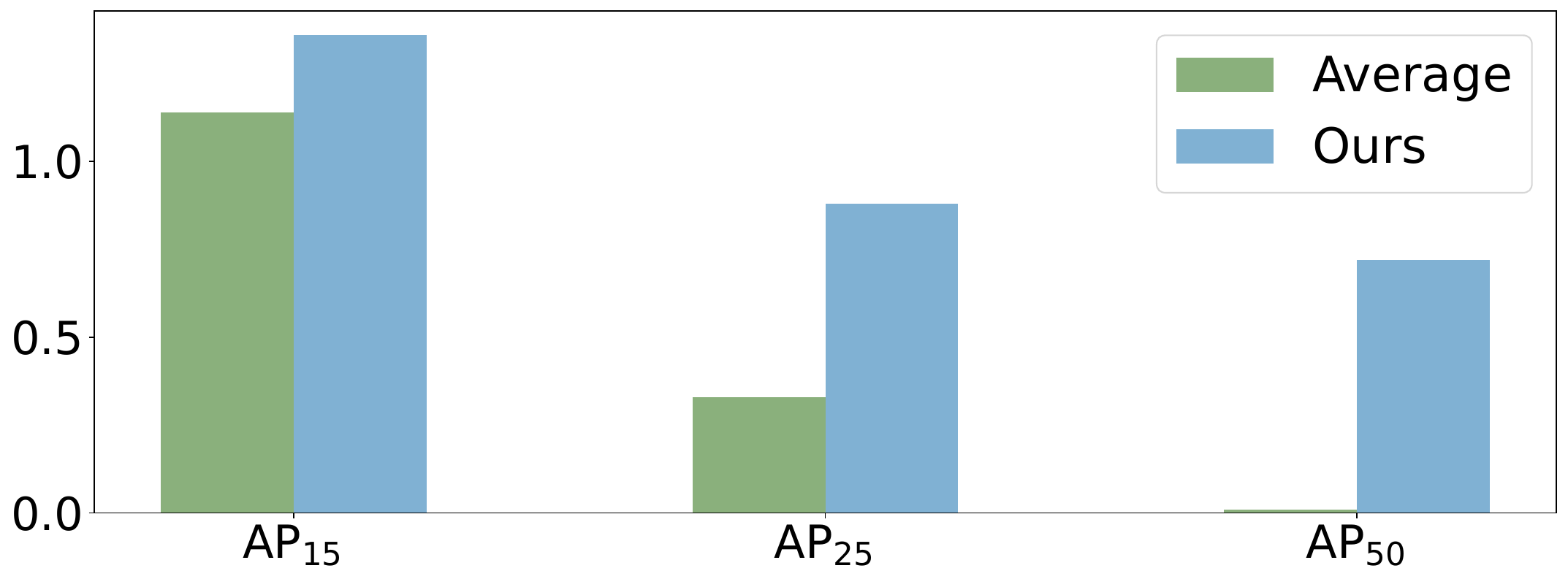} %
    \caption{Ablation studies of different fusion strategies. Improvements compared to the baseline that uses the box with the highest score as the representative are shown. Random optimization of our method is overall superior to other simple strategies.}
    \label{fig:exp-avg}
\end{figure}

Figure~\ref{fig:exp-avg} shows the comparison of different fusion strategies on the ablation sets in Table~\ref{tab:ablation} on CA-1M. The baseline is the strategy that uses the box with the highest score to represent the candidate boxes. The improvements of different strategies compared to the baseline are presented. Additionally, gradient-based fusion strategies using the IoU as the objective loss are overall inferior to the baseline and are not included in the comparison. Our fusion strategy based on particle filtering and random optimization is overall superior to the simple strategy, like averaging. Notably, the improvement of our method compared to averaging is 0.72 in $AP_{50}$, which validates the effectiveness of our method. 

\subsection{Run-time and Memory Analysis}
The core idea of our method is to perform real-time 3D object detection without dense reconstruction. Therefore, it is of great significance to compare our method with other alternatives in terms of running efficiency and memory usage. As shown in Table~\ref{tab:runtime}, our method achieves the best performance in terms of both running time and memory usage. The GPU memory usage of our method is 7.0GB, which is much lower than that of online methods (12.5GB for embodiedSAM~\cite{xu2024embodiedsam} and 21.1GB for OnlineAnySeg~\cite{tang2025onlineanyseg}). Our system FPS is over 20 on average. The main reason for the efficiency of our method is twofold. Firstly, it does not require dense reconstruction, which is a time-consuming process. Instead, it directly detects objects from RGB-D images. Secondly, the proposed random optimization based on particle filtering using pre-sampled swarm templates is highly efficient for online box fusion.

\section{Limitation and Future Work}

Our method is an online reconstruction-free system for 3D object detection, which performs well in static scenes, but is not robust when there are moving objects or humans in the scenario. Moreover, for those extremely cluttered scenarios (e.g., many clothes or toys are stacked together in the closet, or a backpack hangs on the back of a chair), it is quite challenging to detect the bounding boxes of these objects individually. 

\begin{figure}[htb]
    \centering
    \includegraphics[width=0.47\textwidth]{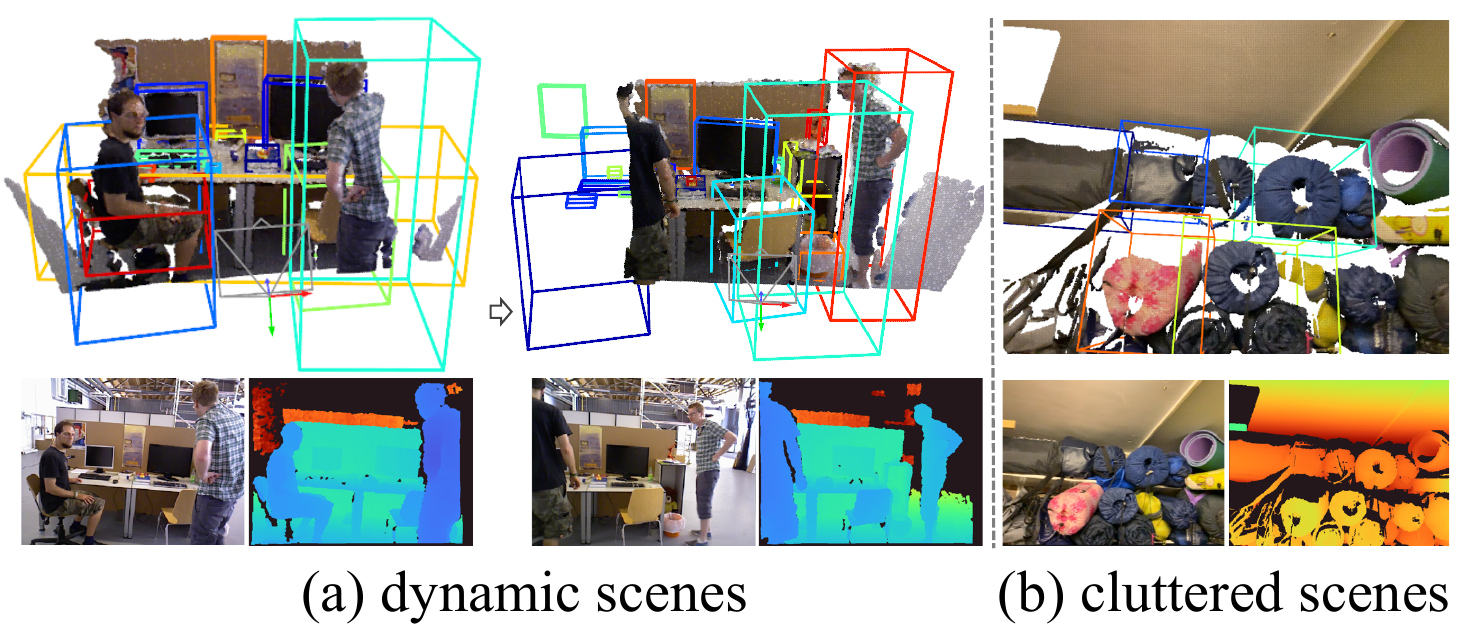} %
    \caption{Typical failure cases. (a) Two humans are walking, which results in redundant and inaccurate boxes. (b) Stacked objects (e.g., bags) in a tiny space cause ambiguity due to their similar appearance. The point cloud is only for visualization here. }
    \label{fig:exp-failurecase}
\end{figure}

Figure~\ref{fig:exp-failurecase} shows the typical failure cases of our method. The left shows the dynamic scene where two humans are walking around the desk, which induces ambiguity due to the dynamic changes and motion blur. The right shows the cluttered scene where there are many bags stacked together in a tiny space, making it hard to detect the boxes correctly and refine the object boxes through multi-view box fusion.

Adapting this reconstruction-free framework to dynamic environments is interesting and promising. Moving objects or humans are common in daily life. Another feasible direction is to leverage the real-time fine-grained 3D object detection to scene graph generation, assisting the embodied agents to have more comprehensive perception about the surrounding environment, which is important for many embodied tasks like navigation or mobile manipulation. 
\section{Conclusions}

In conclusion, we introduce a novel reconstruction-free approach enabling efficient, real-time open-vocabulary 3D object detection. Compared to prior methods leveraging point cloud as the key 3D representation that is computationally expensive, our method employs pre-trained models, Cubify Anything for single-view detection, and CLIP for semantics given streaming RGB-D input. Multi-view predictions are associated via 3D NMS and 2D box correspondence, followed by an IoU-guided random optimization for consistent multi-view box fusion with minimal overhead. Evaluations on CA-1M and ScanNetV2 demonstrate the state-of-the-art performance of our method compared to other online approaches, which validates the novel paradigm for online 3D object detection.
\section*{Acknowledgement}


We kindly appreciate the fruitful discussion and assistance of Yuanhong Yu and Sida Peng. This work is supported in part by the NSFC (62372457), the Major Program of Xiangjiang Laboratory (23XJ01009), NSFC (62325211, 62132021), Young Elite Scientists Sponsorship Program by CAST (2023QNRC001), the Natural Science Foundation of Hunan Province of China (2022RC1104), the National Science and Technology Major Project (2022ZD0115302), NSFC (61379052), the Science Foundation of Ministry of Education of China (2018A02002), the Natural Science Foundation for Distinguished Young Scholars of Hunan Province (14JJ1026). 
{
    \small
    \bibliographystyle{ieeenat_fullname}
    \bibliography{main}
}


\end{document}